# Learning Temporal and Bodily Attention in Protective Movement Behavior Detection

Chongyang Wang[a], Min Peng[b], Temitayo A. Olugbade[a], Nicholas D. Lane[c], Amanda C. De C. Williams[d], Nadia Bianchi-Berthouze[a]

[a]*UCL interaction centre, University College London*, London, United Kingdom

[b]*Chongqing Institute of Green and Intelligent Technology*, *Chinese Academy of Science*, Chongqing, China

[c]*Department of Computer Science*, *University of Oxford*, Oxford, United Kingdom

[d]*Department of Clinical Health*, *University College London*, London, United Kingdom

chongyang.wang.17@ucl.ac.uk, pengmin@cigit.ac.cn, temitayo.olugbade.13@ucl.ac.uk, nicholas.lane@cs.ox.ac.uk, amanda.williams@ucl.ac.uk, nadia.berthouze@ucl.ac.uk

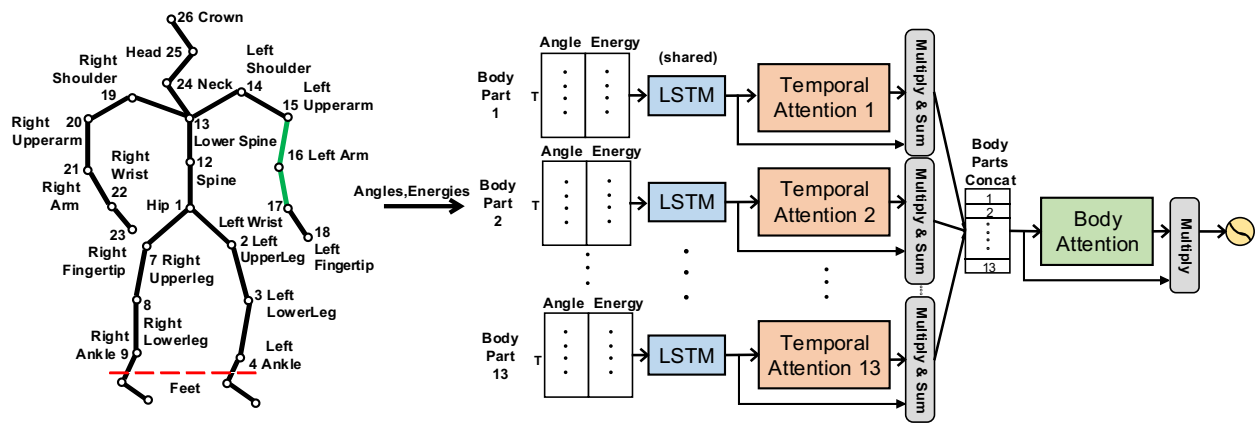

Fig. 1. The overview of the Body Attention Network. Each body part is described by the joint angle plus energy. Data collected from feet were noisy and hence not used in this work.

*Abstract*—For people with chronic pain, the assessment of protective behavior during physical functioning is essential to understand their subjective pain-related experiences (e.g., fear and anxiety toward pain and injury) and how they deal with such experiences (avoidance or reliance on specific body joints), with the ultimate goal of guiding intervention. Advances in deep learning (DL) can enable the development of such intervention. Using the EmoPain MoCap dataset, we investigate how attention-based DL architectures can be used to improve the detection of protective behavior by capturing the most informative temporal and body configurational cues characterizing specific movements and the strategies used to perform them. We propose an end-to-end deep learning architecture named BodyAttentionNet (BANet). BANet is designed to learn temporal and bodily parts that are more informative to the detection of protective behavior. The approach addresses the variety of ways people execute a movement (including healthy people) independently of the type of movement analyzed. Through extensive comparison experiments with other state-of-the-art machine learning techniques used with motion capture data, we show statistically significant improvements achieved by using these attention mechanisms. In addition, the BANet architecture requires a much lower number of parameters than the state of the art for comparable if not higher performances.

## I. INTRODUCTION

Physical rehabilitation is important for the management of chronic pain (CP) [4][5][8]. Fear of pain and/or injury in people with CP results in reducing physical activity or using movement strategies (e.g., guarding, stiffness, hesitation, bracing) [1][2][40], collectively called protective behaviors [3]. Such behaviors cause further debilitation and reduced participation in valued activities, e.g. employment or social life [2][4], and so are important targets for intervention. For example, in clinical settings, physiotherapists modify psychological support, feedback, and exercise movement to change CP patients' specific movement behavior fear [7][8].

As physical rehabilitation for CP moves from clinical settings to self-management, ubiquitous technology is targeted as a tool for providing support in lieu of a physiotherapist's affect-based personalized support [9]. In order for technology to do so, it is essential that it automatically assesses movement and detects protective behavior. Studies of automatic analysis of emotion-influenced movement behavior are rare, as most studies in pain scenario focus on facial expression of pain or physiological responses to acute/stimulated pain [10][11]. This is partly due to scarce data and possibly limited appreciation, in the computing field, of the importance of bodily behavior over facial expression in providing information about a person's psychological capability to manage his/her condition [5][12].

To drive the research on protective behavior, the EmoPain dataset [6] was created. The dataset includes full-body motion capture (MoCap) data for 26 anatomical joints and surface electromyography (sEMG) data collected from 4 locations on the back. Participants were 22 people with chronic lower back pain (CLBP) and 26 healthy people. Recorded activities included: One-leg-stand, Stand-to-sit, Sit-to-stand, Reach-forward and Bend - typical everyday activities that are generally challenging for those with CLBP [7][9]. Previous studies based on Mocap and sEMG data of the EmoPain dataset mainly employed vanilla neural networks [15] and feature engineering approaches [6][13][14], where the dynamic biomechanics of movements are only used to guide feature design. Unlike acute rehabilitation where a gold-standard movement trajectory (and its deviation) informs intervention, in chronic pain, fear of injury, fear of pain, and anxiety lead the person to engage body parts in ways that are not biomechanically necessary or efficient but may increase sense of control and reduce fear. Thus, in this paper we present an end-to-end neural network architecture called BANet that can, across different types of movement, self-learn when (temporal attention) and what (bodily attention) subsets of the anatomical joints contribute most to the detection of protective behavior. With the analysis of the learned attention scores, we explore how the various behaviors observed lead to the network shifting attention to different body parts as necessary.

To avoid ambiguity, we clarify the following terms used in this paper: 'sample' refers to a single data vector at each single timestep (1/60 of a second as the sensor captured at 60Hz); 'segment' or 'frame' refers to a small data window containing several samples within a data instance; 'instance' refers to the full data sequence containing all the activities (sit-to-stand, bend, etc.) performed by a subject during one trial (each participants underwent two trials of two difficulty levels, for details see [6]). Given that there are five types of protective behaviors (hesitation, guarding, stiffness, bracing and support) [6] but limited volume of the full dataset, we use the experience learned from [15] about treating all five behaviors as one single category called *protective behavior*. The input of BANet are local joint angles and their energies (the square of the angular velocity), with each angle calculated from three relevant anatomical points (e.g., the knee angle formed by joining joints on the neck, knee and ankle) as shown in Fig. 3 (bottom-right). The use of these as input is based on previous studies [6][27]. The contribution of the work in this work can be summarized as:

- We propose a novel deep learning architecture performing spontaneous temporal and bodily part subset selection. Here, we focus only on MoCap data particularly comprising streams of joint angle, but the network can be easily adapted to data of joint positions or even data collected from multiple sensor types (e.g., MoCap plus EMG data)

- Through a range of experiments on the EmoPain dataset, we demonstrate that our method can achieve state-of-the-art results, if not slightly higher, with fewer trainable parameters for the detection of protective behavior.

- With attention score visualization and analysis, we discuss how such mechanisms could help better understanding of protective behavior from real-life measurements, rather than just lab-based observations.

## II. RELATED WORKS

In this section, we first present the state-of-the-art on the automatic detection of protective behavior in chronic pain. Then we review studies on attention mechanisms, especially those on human activity recognition (HAR) using wearable sensors.

### A. Automatic Analysis of Protective Behavior

An earlier study on the automatic detection of protective behavior was conducted by Aung et al. [6], using Random Forest (RF), a traditional machine learning algorithm [16], on features extracted from body movement data (the EmoPain dataset). These features were the range of 13 joint angles (based on location of incidental points and two immediate neighbors), the mean of corresponding energies (defined as the square of angular velocity), and the mean of sEMG data. These were used to predict protective (movement) behavior in each activity instance, with mean square error between 0.019 to 0.034 (mean = 0.44, standard deviation = 0.16) for different movement types.

A more recent study by Wang et al. [15] explored the use of recurrent neural networks on the joint angles, energies and sEMG data from [6]. Their stacked-LSTM architecture produced higher performance than other vanilla deep learning architectures (e.g., convolutional neural network) in detecting protective behavior over sliding windows across different activity types, rather than having to build movement-dependent models as in [6]. To address the higher data demand typically required by neural network algorithms, data augmentation was carried out on training segments by adding Gaussian noise and randomly discarding data samples, which lead to an F1-score of 0.815, a level appreciably better than using the original dataset. Experimentation also showed that 3 seconds with zero-padding was the optimal sliding-window setting parameter. Their study suggests the applicability of LSTM networks for the detection of protective behavior, based on MoCap and sEMG data. An obvious limitation of both [6] and [15] is that they considered all body-parts with equal importance across time and activity whereas protective behavior may occur in a specific stage of the activity and involve only a specific set of body parts, possibly different stages and body parts across activity types and across the CP population. Hence, by processing the full-body MoCap (and sEMG) data in a traversal manner, redundant information and less informative data are retained, possibly constituting noise and thus reducing the performance of the model. In our work, we aim to let the architecture focus on the relevant configuration of protective behavior by integrating attention mechanisms in the LSTM-based network. We exploit MoCap data alone for two reasons. First, our aim is to better understand how the model shifts between different body areas that are easily visually understood and compare them with physiotherapists' observations, which do not rely on sEMG during typical consultations. Second, we are keen to understand what such modality alone (without sEMG) can achieve using attention mechanisms with respect to the state of the art. Still, as previous studies [15] [33] have shown that sEMG data are critical for high performance, future work should explore the combination of the two types of data.

### B. Attention Mechanism Adapted for HAR on MoCap data

A common scenario in HAR is the acquisition of data with MoCap sensors (e.g., gyroscope and accelerometer) attached to different body locations. Targeting the spatial and temporal subset selection within such data, attention mechanisms have been recently explored to improve HAR performances on MoCap data. To enable understanding of the relevance of each sensor in such scenario, Zeng et al. [19] proposed an attention-based LSTM framework, where a sensor attention module was used at the input level and for each timestep, with an additional temporal attention module at a later layer. Their sensor attention module was implemented with input from different sensors at single timestep, while temporal attention was computed based on the output of the LSTM layer. This improved the performance on three HAR benchmarks (PAMAP2 [22], DG [23] and Skoda [24]). Their visualization of the attention scores satisfied the expectation of subset learning of sensors at important moments. Along with the same route, Murahari et al. [20] focused on temporal attention, which they embedded at the end of a convolutional LSTM network (Conv-LSTM) [25]. They used tanh and softmax functions to compute the attention scores with the LSTM outputs, the weighted sum of all previous LSTM hidden states instead of only the last hidden state, was used for classification. Another study, by Yao et al. [21], was motivated by the problem of sensor reliability in mobile sensing, where multiple sensors are deployed at same time but only a hidden subset provides reliable information. They adapted the DeepSense framework [26] by additionally using two softmax functions to compute the attention scores specific to sensors, at a lower layer, and temporal attention, at a higher layer. This led to better performance on the HAR dataset [32] compared with results achieved with the original DeepSense framework.

These works suggest that explicitly designing an attention mechanism can help a neural network better learn patterns in data from multiple sources (e.g., multiple bodily parts). However, we noticed two main limitations: i) sensor attention and temporal attention are computed based on different scales of information, i.e., the computation of sensor attention with the low-level input data at each single timestep and temporal attention with the output of LSTM/GRU layer across a prior of time spanning multiple timesteps, which created a gap between the two attention results making it inappropriate to combine them together directly; ii) computing sensor attention per timestep is impractical as the time then attended to is too local. We argue that the relevance of a sensor dimension (each joint angle in our case) can only be understood over a time period (i.e. over a movement segment), rather than at a single timesteps. Further, where in HAR, activities can perhaps be recognized based on temporally-local relationships, protective behavior detection is a higher-level analysis that may thus require a longer period of perception before a clear judgment can be made. Also, as the size of the dataset for protective behavior is much smaller than typical HAR datasets, it is valuable for more care to be taken in designing an efficient learning network.

## III. METHODLOGY

In this section, the BANet architecture is first presented. Then we describe the attention mechanisms designed considering the characteristic of protective behavior and targeting the issues we found in previous literature. We also present the data preparation approach including segmentation and augmentation, as well as important experimental settings.

### A. The BANet Architecture

An overview of BANet is given in Fig. 1. The input to the BANet is a $2 \times 13$ low-level movement matrix (angles and energies from 13 body parts), for each sample/timestep in a movement segment. A shared vanilla LSTM subnetwork is used to extract the temporal information independently for each of the 13 body parts, i.e. given a data segment, the output of such temporal decoder is a matrix of hidden states $H_t^{c,k} = [h_1^{c,k}, ..., h_t^{c,k}]$ with $h_t^{c,k} = [h_t^{c,1}, ..., h_t^{c,k}]'$, where $C \in$ {1,2,...,13} for the 13 body parts, $t = 1,2,...,T$ where $T$ is the temporal length of input data matrix, $k = 1,2,...,K$ where $K$ is the number of hidden units used in the temporal encoder.

### B. Temporal and Bodily Attention Learning

The attention mechanism of BANet is implemented with two stages: a temporal attention module, separate for each body part, followed at a higher-level by a global and bodily attention module. As we can see in Fig. 1, the computation of temporal and bodily attention is on same information source, the temporal information extracted with LSTM network from each body part. The two attention modules are described below:

*1) Temporal Attention Module:* To learn the attention score $a_t^C$ for $H_t^{C,K}$, we use a $1 \times 1$ convolutional layer followed by a softmax layer:

$$a_t^C = softmax(W_\alpha * H_t^{C,K}) \quad (1)$$

$$softmax(x_i) = \frac{\exp(x_i)}{\sum_{i=1}^I \exp(x_i)} \quad (2)$$

where $W_\alpha$ is a learnable weight matrix, and $*$ is convolution operation. Fig. 2 (above) illustrates this such computation. Unlike a the fully-connected layer, the $1 \times 1$ convolution layer acts as a linear embedding which limits irrelevant connections among the input matrix (in our case is the temporal connection of samples at within a frame). Thus, the $1 \times 1$ convolution layer can help to minimize the number of trainable parameters. We experiment with the variant using fully connected layer to justify this. The temporal attention further includes a merge of the attention score with the original output of the temporal decoder through a multiplication followed by a sum-up over samples:

$$H_C^K = \sum_{t=1}^T a_t^C H_t^{C,K} \quad (3)$$

The output of the temporal attention module is a matrix of the weighted-sum of the temporal information from the 13 body parts, which can be written as $H_C^K = [(h_1^1, ..., h_1^K)', ..., (h_{13}^1, ..., h_{13}^K)']$.

*2) Bodily Attention Module:* So far, the network has processed the data segment separately for each body part. A global attention score for each body part is now computed, based on the output from temporal attention layer. This is in order to learn the subset of the body parts that play a key role in the detection of protective behavior during a given movement segment. We use two fully-connected layers with tanh and softmax activations to compute the bodily attention score $b_c$:

$$b_c = softmax\left(tanh(W_\beta H_c^K)\right) \quad (4)$$

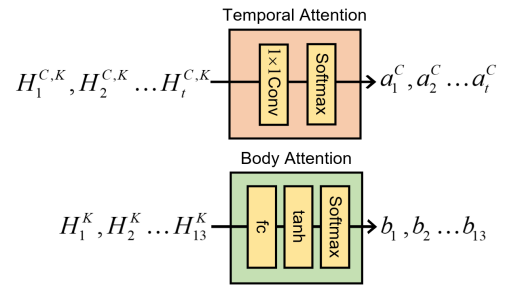

Fig. 2. Above: temporal attention block; Below: the body attention block.

where $W_\beta$ is a learnable weight matrix. The body attention module is completed by merging the body attention score with the original output of the temporal attention module:

$$attenH_C^K = b_c \odot H_C^K \quad (5)$$

Such **attention-over-attention** structure of BANet finally produces a K × 13 matrix $attenH_C^K$ which encodes the importance of each body parts at important moments (samples) for the input segment. With such output, the detection is finally completed with a fully-connected layer using softmax activation.

*C. Data Preparation and Experimental Settings*

*1) Data Segmentation and Augmentation:* As introduced earlier, we use only the motion capture data of the EmoPain dataset collected from 18 CLBP patients and 12 healthy people performing functional activities [6]. In total, there are 46 activity instances where each instance is around 10 minutes long and contains sequences of sit-to-stand, stand-to-sit, bending, reaching forward and one-leg-stand activities. Such limited data size is typical when real data (as opposed to acted) is collected from patient cohorts. Following the approach in [6], each sample is characterized by 13 full-body joint-angles as well as the energies of these. Each joint angle is formed by connecting three body-joints in the 3D Cartesian space, and have the advantage, over joint positions, of being invariant to the translation and change of the reference frame [28]. The energy is the square of the respective angular velocities. The description of the 13 joint-angles is shown in Fig. 3 (bottom-right).

To create the training and test data for our experiments, we run a sliding window, length = 3 seconds and overlapping ratio = 75% based on findings in [15], within each activity type in the data instance. Zero-padding is used when the window slides beyond the end of a given activity type. This amounts to a total of 2,569 segments. The ground truth for each segment is set based on majority-voting, where a segment is labelled with *protective behavior* if at least 50% of the samples within it had been rated as protective by at least 2 out of the 4 raters, and *non-protective behavior* otherwise.

For the training of BANet and of other architectures evaluated for comparison, we apply two augmentation techniques, both previously used in [15][39]. The first technique is based on creating new instances by adding normalized Gaussian noise to the original data with 3 different standard deviations: 0.05, 0.1 and 0.15. The second approach creates new instances by randomly altering (set to 0) the samples across time and body parts, with selection probability of 0.05, 0.1 and 0.15. This latter method aims to simulate the presence of incomplete data. The use of the two approaches leads to 18,653 segments, where 11,373 segments are labelled as non-protective (from both healthy participants and patients) and 7,280 segments are labelled as protective (only from patients).

*2) Experimental Settings:* The BANet is implemented with Keras and TensorFlow. For the LSTM network acting as temporal information encoder, we use a 3-layer LSTM network with 8 hidden units in each layer. Dropout with probability of 0.5 is used after each LSTM layer. In the full network, weights are updated with Adam optimizer [31]; a learning rate of 0.003 and batch size of 40 are applied.

The validation method used in this study is the standard leave-one-subject-out (LOSO) cross-validation, to test the generalization ability of a model to unseen subjects. In our work, the detection of protective behavior is a binary task as we merged the 5 categories of protective behaviors [6] [15] in the EmoPain dataset into one. We report the mean F1 score as the metric. Statistical tests particularly repeated-measures ANOVA and post-hoc paired t-tests are used to compare the performances of different architectures.

IV. EXPERIMENT RESULTS AND DISCUSSION

In this section, we first report the results of the comparison experiments. Then, we visually analyze and discuss the movement cues of protective behavior that emerge from the attention mechanisms.

*A. Comparison Experiments*

We compare BANet with vanilla neural networks used in earlier studies: i) Convolutional LSTM (Conv-LSTM) [29], with convolution kernel size of 1 × 10, max pooling size of 1 × 2, 10 filters, 28 LSTM hidden units and batch size of 50; ii) Bi-directional LSTM (bi-LSTM), with 14 LSTM hidden units followed by a Dropout with probability of 0.5, and batch size set to 40; iii) stacked-LSTM, a vanilla 3-layer LSTM network [15] with each layer of 28 hidden units followed by Dropout with probability of 0.5, the batch size is set to 20. For all the neural networks, the hyperparameters were tuned through grid-search, and the Adam optimizer [31] is used for with learning rate of 0.003.

We also compare with a variant of the BANet with a fully-connected layer used in the temporal attention computation (BANet-dense) instead of a 1 × 1 convolution layer. In addition, we compare our work with the approach used in related HAR studies [19] [20] [21], where the sensor attention was computed before the extraction of temporal information. As such, we create a variant (BANet-compat, for BANet compatibility version) where the computation of body attention was done at input level instead of at feature fusion level with the same attention algorithms presented in last section. For BANet-compat, at each timestep, the body attention scores were computed for the 13 body parts and after multiplication, all the timesteps are concatenated for the temporal information extraction and temporal attention computation that follow. Finally, the output to be classified has the same size of $k × 13$ as the BANet ($k$ is the number of hidden units of the LSTM encoder). Additionally, to show the impact of the two attention modules introduced, we provide the results of BANet-body where only the body attention is implemented and BANet-time where only the temporal attention is computed. The same hyperparameters and training strategies used with BANet were employed.

Results for the comparison experiments are shown in Table I. As shown, the proposed BANet achieved the best results (accuracy of 0.8688, mean F1-score of 0.844), with a smaller parameter size of 2,131 in comparison to other tested LSTM-based architectures (parameter sizes ranging from 14,000 to 40,000). The parameter reduction is obtained in BANet through the use of: i) the temporal information extraction strategy independent of body parts, providing data with a smaller dimension and allowing the LSTM layer within BANet to have a smaller number of hidden units; ii) a 1 × 1 convolution instead of fully-connected operation for computing the temporal attention, the former being a critical advantage due to the many timesteps (180 timesteps) of the input to this layer. The second best is achieved with BANet-body which shows the importance of focusing on a subset of joints-angle (rather than all) for the detection of protective behavior. Instead, the BANet-time that only learns separately the temporal attention separately for each joint does not achieve high accuracy results. This is expected and is due to the lack, in this network, of global processing over all body parts. The next best result is achieved by the stacked-LSTM [15] (accuracy of 0.8534, mean F1-score of 0.812). Although the result is very similar to BANet's (see also their confusion matrices in Table II), stacked-LSTM requires a much larger number of parameters (18,986). On the other hand, except for that the BANet-compat is only a representative of the architectures used in [19][20][21], the results imply that encoding the importance of body joints at a single timestep is not valuable to the detection of protective behavior, but should be delayed to a higher-level processing given a period of data input, as in the BANet.

TABLE I. RESULTS FOR OUR BANET AND COMPARISON ALGORITHMS

| Architecture (Bold: best performance) | | Accuracy | Mean F1 | p-value | Parameters Size |
|---|---|---|---|---|---|
| Comparison architecture | Conv-LSTM [29] | 0.8059 | 0.737 | 0.049 | 40,940 |
| | bi-LSTM [30] | 0.8460 | 0.804 | 0.05 | 14,282 |
| | stacked-LSTM [15] | 0.8534 | 0.812 | 0.026 | 18,986 |
| Proposed architecture | BANet-compat | 0.6630 | 0.572 | 0.0001 | 6,204 |
| | BANet-dense | 0.8167 | 0.789 | 0.019 | 65,430 |
| | BANet-time | 0.806 | 0.758 | 0.09 | 1,767 |
| | BANet-body | 0.867 | 0.831 | 0.167 | 2,023 |
| | **BANet** | **0.8688** | **0.844** | - | 2,131 |

TABLE II. CONFUSION MATRICES FOR BANET AND STACKED-LSTM

| BANet | | Non-Protective | Protective |
|---|---|---|---|
| Groundtruth | Non-protective | 1491 (92.84%) | 115 (7.16%) |
| | Protective | 331 (31.83%) | 709 (68.17%) |
| stacked-LSTM | | Non-Protective | Protective |
| Groundtruth | Non-protective | 1451 (90.35%) | 155 (9.65%) |
| | Protective | 322 (30.96%) | 718 (69.04%) |

A repeated-measure ANOVA shows statistical-significant differences in the performances across folds between the proposed BANet and the others: $F(3.072, 89.099)=15.612$, $p=<0.0001$, $\mu^2 = 0.519$). Post-hoc paired t-test with Bonferroni correction shows that significance does not hold only for the BANet-body (p=0.167). When the BANet-body itself is compared with the other architectures, significant difference is found only for the BANet-time (p=0.022). This suggests that the impact introduced by the body attention module is more significant than temporal attention whereas the combination of the two attention mechanisms consistently leads to even better results.

*B. Analysis on Attention Scores*

While a full analysis of attention scores is out of the scope of this paper due to space limit, we highlight some of the trends to enable understanding to what extent the two attention mechanisms capture aspects of protective movement strategies. Besides improving classification performance, such scores may help further understanding of pain behavior, which is still predominantly based on observations rather than objective measurements. Fig. 3 (top) shows boxplots of the joint attention scores for all test segments over all folds organized by movement type. It is interesting that boxplots for healthy participants (green) are quite narrow compared to those of participants with CLBP (blue and orange). This can be related back to the literature in two ways.

First, even when people were not asked to perform a movement according to ideal trajectories [34], healthy subjects tend to perform simple everyday movements in a quite similar way. Only a few healthy participants' boxplots are slightly wider, especially in the bend and reach-forward movements that may be more biomechanically demanding. Instead, pain literatures (e.g., [38]) show that people with CLBP do tend to show a wide variety of strategies when performing simple movements, this may depend on what part of the body they perceive as vulnerable (e.g., in stand-to-sit, avoiding bending the trunk by bending the legs more such as in Fig. 3 (bottom)-P17, or reducing weight on the legs by twisting the trunk such as in Fig. 3 (bottom)-P14.

Second, the limited width of the healthy participants' boxplots supports the fact that each joint maintains fairly constant relevance throughout the movement, highlighting synergetic use of engaged body parts in performing the movement [6] [35]. See, for example, the temporal attention scores of control participant C16 in the stand-to-sit (Fig. 3-bottom). On the other hand, such ability of characterizing the healthy people may account for the better true-negative rate of BANet seen in Table II. However, this constancy is not the case for CLBP participants, as the wider boxplots suggest. As discussed in [33] [35] [36], people suffering from CLBP tend to engage different body parts at different stages of the movement rather than in a synergetic way, despite the fact that such strategies make the movement more difficult to execute. This approach could be due to perceived higher control over the movement, and/or the fear of pain or injury, or reduced proprioception/coordination capabilities as discussed in [33] [38]. For example, let us analyze in more detail P14's darker (and so higher) temporal attention scores during stand-to-sit (Fig. 3-bottom). P14's engagement of the leg and shoulder at the initiation of the sit-down suggests hesitation (as indicated by physiotherapists in [6] [37]). We know from video analysis that this initial hesitation is followed by a horizontal twist of the shoulder (which is captured by the right shoulder's score) followed by the bending of the neck to check for the chair position, then still the twisting of the shoulder (captured by the left shoulder score) to use the arm (left elbow bent beside the trunk) for support on the chair to minimize the load on trunk and on legs. Healthy participant C16, the participant also uses the arms, but behind the body (rather than on the side) and these reach for the chair together with the trunk, and not used as support for legs or back.

V. CONCLUSION

This paper investigates the use of both temporal and bodily attention mechanisms combined LSTM layers to improve the quality of detection of protective behavior in people with CLBP. The study is based on the EmoPain MoCap data. In comparison to the state of the art, our architecture delays the attention processes to the second and third layer of the architecture to enable primary learning of low-level features as the movement is processed. In doing so, both attention mechanisms work on a higher-level representation of the movement. The results show that such an approach leads to a substantial improvement (F1 from 0.572 to 0.844). Further, it shows results slightly higher than other LSTM-based architectures with a critical decrease in number of hyper-parameters (from 40,940 to 2,131). The results also suggest that bodily attention mechanism plays a more important role than the temporal attention mechanisms (F1 of 0.831 vs 0.758 respectively), still the combination of the two mechanisms leads to much higher performances (F1 of 0.844). This suggests that the temporal attention mechanism may capture more detailed local temporal dynamics missed by the bodily attention. In addition, such temporal dynamics may be also critical in discriminating between strategies. The paper concludes with discussion of some examples of how the two types of attention mechanism scores capture aspects of protective movement, suggesting that such systems, deployed in everyday rehabilitation activity, could be used not only to provide a more effective therapy but to also contribute to the pain literature by enabling a better understanding of protective behavior in everyday life.

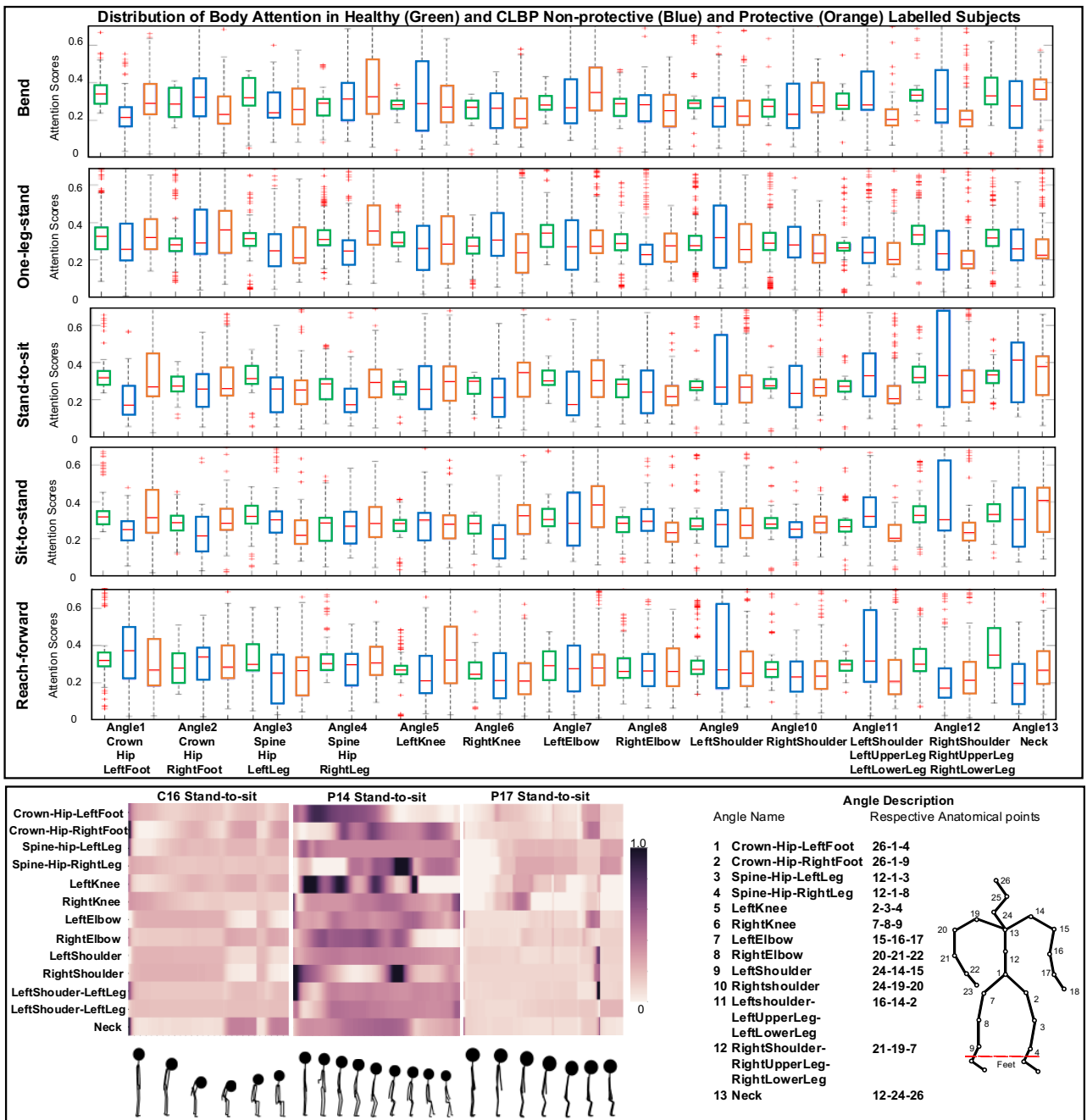

Fig. 3. Above: Boxplots for the distribution of body attention computed by BANet for each testing data organized by movement type. Bottom-Left: Temporal attention map from BANet for testing instance of patient number 14, 17 and healthy subject number 16 with their respective MoCap data (stick figures). Bottom-Right: Angle description for the 13 joint angles used in this paper.

ACKNOWLEDGMENT

The project was supported by the Future and Emerging Technologies (FET) Proactive Programme H2020-EU.1.2.2 (Grant agreement 824160; EnTimeMent) and Chongyang Wang is supported by the UCL Overseas Research Scholarship (ORS) and Graduate Research Scholarship (GRS).